\documentclass{elsarticle}
\usepackage{amsmath, amssymb,bm}
\usepackage{booktabs}
\usepackage{graphicx,epstopdf,subfigure}
\usepackage{caption}
\usepackage{multirow}
\usepackage{tabularx,threeparttable}
\usepackage{array}
\usepackage{comment}
\usepackage{mathtools}
\usepackage[hidelinks]{hyperref}
\usepackage{xcolor}
\usepackage{url}

\usepackage{enumitem}
\usepackage{bbm}
\usepackage{bm}
\usepackage[ruled]{algorithm2e}

\journal{Medical Image Analysis}

\biboptions{authoryear}

\begin{document}

\begin{frontmatter}

\title{Multi-Task Recurrent Convolutional Network with Correlation Loss for Surgical Video Analysis}

\author[firstadd]{Yueming Jin}
\author[firstadd]{Huaxia Li}
\author[firstadd]{Qi Dou\corref{cor1}}
\cortext[cor1]{Corresponding author: qdou@cse.cuhk.edu.hk (Qi Dou).}
\author[firstadd,senadd]{Hao Chen}
\author[thirdadd]{Jing~Qin}
\author[firstadd]{Chi-Wing~Fu}
\author[firstadd]{Pheng-Ann Heng}
\address[firstadd]{Department of Computer Science and Engineering, The Chinese University of Hong Kong}
\address[senadd]{Imsight Medical Technology, Co, Ltd, China}
\address[thirdadd]{Centre for Smart Health, School of Nursing, The Hong Kong Polytechnic University}

\begin{abstract}
Surgical tool presence detection and surgical phase recognition are two fundamental yet challenging tasks in surgical video analysis and also very essential components in various applications in modern operating rooms. While these two analysis tasks are highly correlated in clinical practice as the surgical process is well-defined, most previous methods tackled them separately, without making full use of their relatedness. In this paper, we present a novel method by developing a multi-task recurrent convolutional network with correlation loss (\emph{MTRCNet-CL}) to exploit their relatedness to simultaneously boost the performance of both tasks. Specifically, our proposed MTRCNet-CL model has an end-to-end architecture with two branches, which share earlier feature encoders to extract general visual features while holding respective higher layers targeting for specific tasks. Given that temporal information is crucial for phase recognition, long-short term memory (LSTM) is explored to model the sequential dependencies in the phase recognition branch. More importantly, a novel and effective correlation loss is designed to model the relatedness between tool presence and phase identification of each video frame, by minimizing the divergence of predictions from the two branches. Mutually leveraging both low-level feature sharing and high-level prediction correlating, our MTRCNet-CL method can encourage the interactions between the two tasks to a large extent, and hence can bring about benefits to each other. Extensive experiments on a large surgical video dataset (Cholec80) demonstrate outstanding performance of our proposed method, consistently exceeding the state-of-the-art methods by a large margin (e.g., $89.1\% ~v.s.~ 81.0\%$ for the mAP in tool presence detection and $87.4\%~v.s.~84.5\%$ for F1 score in phase recognition). The code can be found on our project website.

\end{abstract}

\begin{keyword}
Surgical video analysis, multi-task learning, correlation loss, deep learning.
\end{keyword}

\end{frontmatter}


\section{Introduction}

With the advancements in medicine and information technologies,
the operating room (OR) has undergone tremendous transformations evolving into a highly complicated and technologically rich environment (\cite{cleary2005or,james2007eye,lalys2014surgical,bouget2017vision}).
These transformations allow the execution of more complex procedures and also increase the amount of information available in modern OR.
To better tackle this new OR scenario, the context-aware systems (CAS) have gradually been developed to provide detailed comprehension of rich information and contextual support to the clinicians (\cite{bricon2007context,dergachyova2016automatic}).
With interpreting the operation procedure and tool usage, automated surgical phase recognition and tool presence detection serve as the primary functions in CAS and such accurate systems are expected to be highly demanded (\cite{padoy2012statistical,lalys2014surgical,wesierski2018instrument}).

Specifically, automatically recognizing the surgical phase and tool enables CAS to assist clinicians during two periods: intra-operation and post-operation.
The intra-operative recognition is able to generate real-time warning for clinicians by detecting rare cases and unexpected variations~(\cite{bouget2015detecting}), and to support decision making of junior surgeons through timely communication within surgical team (\cite{quellec2014real,quellec2015real,forestier2015automatic}).
The online recognition can also help to improve OR resource management. 
By knowing which surgical workflow is currently occurring and which tool is utilized, the completion time of surgery can be estimated. 
Therefore, it can facilitate relevant clinical staff to prepare the following patient in advance, resulting in minimal patient waiting time and maximal OR throughput~(\cite{twinanda2017endonet,bouget2017vision}).
In addition, the post-operative recognition can enhance the efficiency of surgical report documentation and video database indexing, which are currently tedious and time-consuming manual jobs.
The indexed record of surgical procedure can further facilitate the surgeon training, review and skill assessment (\cite{zappella2013surgical, ahmidi2017dataset,sarikaya2017detection}).
The fully annotated database can also be utilized to generate the statistical information, which is beneficial for the surgical workflow optimization (\cite{bhatia2007real,wesierski2018instrument}).

However, developing automated methods to recognize tool presence and surgical phase from surgical videos is very challenging.
First, there is a large variety of surgical tools with some abnormal cases, such as partial appearances and overlap of multiple tools.
Second, complicated surgical scenes lead to limited inter-phase variance while high intra-phase variance. 
Lastly, observed surgical scenes are often blurred due to the camera motion and gas generated by tools, and even completely occlude when blood stains the camera lens.
Extra noise and artifacts introduced by consequent lens cleaning process make the recognition tasks even harder.
To meet these challenges, early methods relied on handcrafted features such as gradient magnitude~(\cite{blum2010modeling}), combinative descriptors~(\cite{lalys2012framework}), and intensity values~(\cite{zappella2013surgical}). 
However, the empirical design of these low-level features heavily depends on domain knowledge and would be insufficient to represent the complicated characteristics of surgical videos.
With the revolution of deep learning, many attempts of adapting convolutional neural networks (CNNs) and recurrent neural networks (RNNs) on surgical video analysis have been explored and achieved promising performance~(\cite{dipietro2016recognizing,sahu2017addressing,jin2018sv}). 
Unfortunately, most existing deep learning based methods addressed the tool and phase recognition tasks independently, without considering the intrinsic association between them.

\begin{figure*}[t]
	\centering
	\includegraphics[width=1\textwidth]{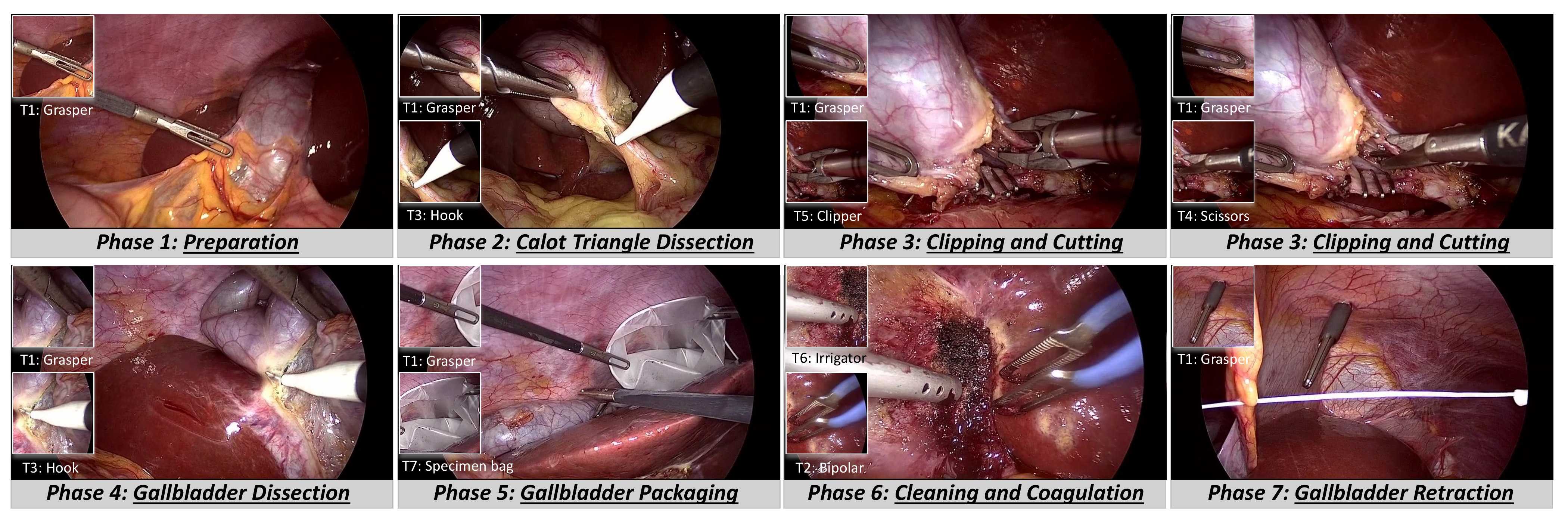}
	\vspace{-1mm}
	\caption{Illustration of definition and correlation of each phase and tool presence in surgical videos, taking the cholecystectomy procedure as an example.}
	\label{fig:relation}
\end{figure*}

According to the regulation of surgery procedure, surgeons are requested to perform specified operations with corresponding sets of instruments for different surgery phases.
Therefore, there exists a high correlation between the surgical phase and tool usage. 
Taking the cholecystectomy procedure as an example (see Figure~\ref{fig:relation}), hooks are often used to perform the dissection operations; clipper and scissors are required in clipping and cutting stage.
In fact, many previous works directly employed binary instrument usage signals to perform phase recognition, which manifested the benefit of tool information to phase recognition (\cite{padoy2012statistical,forestier2015automatic}).
Recently, \cite{twinanda2017endonet} implemented a multi-task framework with shared early layers and incorporated tool information in the feature learning process, which firstly achieved joint tool and phase recognitions.
The promising performance demonstrates that effectively leveraging such relatedness plays an essential role in improving both tasks, i.e., tool presence detection and phase recognition.

The correlation between the multiple tasks is often quite complicated. For example, in surgical videos, the same tool may present in different phases, while an operation phase may involve a variety of instrument combinations. 
To this end, the shortcomings existing in the aforementioned approaches may fail to precisely capture the correlation.
For example, the method proposed by \cite{twinanda2017endonet} uses hidden Markov model (HMM) to enforce the temporal constraints on the phase prediction, instead of introducing sequential information in the network training procedure, which plays a key factor in tackling video-based tasks.
Therefore, how to precisely capture and sufficiently leverage the close yet subtle correlations between these two tasks remains a problem to be further investigated.

In this paper, we present a novel method, i.e., multi-task recurrent convolutional network with correlation loss (\emph{MTRCNet-CL}), to simultaneously tackle tool presence detection and phase recognition tasks.
The proposed end-to-end framework is capable of comprehensively alleviating the shortcomings of other surgical video analysis approaches and  improve the ability of correlation capture. 
The source codes and relevant supporting documents can be found on our project website\footnote{\url{https://github.com/YuemingJin/MTRCNet-CL}}.
Our main contributions are summarized as follows:

\begin{itemize}
	\item
	We develop a novel multi-task recurrent convolutional network for surgical video analysis. It consists of two branches, which share early convolutional layers, and are well designed for solving particular tasks at respective higher layers.
	Given that temporal information is crucial for phase recognition, we employ a recurrent unit, i.e. long-short term memory (LSTM), in this branch to encode sequential dependencies. In this way, we form a multi-task learning model composing of CNN and RNN modules.
	
	\item 
	We propose a correlation loss to provide additional regularization inspired from domain knowledge, by minimizing the divergence of probability predictions for the two tasks.
	This mechanism is to penalize the model, when the correlated tasks resulting in conflicting predictions.
	The features which are discriminative for each task can have dynamic interactions at hierarchy during the training process, presenting a novel style of multi-task learning.
	
	\item 
	We extensively validate our proposed MTRCNet-CL on a large endoscopy surgical video dataset (Cholec80). Our method outperforms existing state-of-the-art approaches significantly and consistently (e.g., $89.1\% ~v.s.~ 81.0\%$ for the mAP in tool presence detection and $87.4\%~v.s.~84.5\%$ for F1 score in phase recognition), demonstrating outstanding efficacy of our developed multi-task learning strategy for surgical video analysis. The source code and relevant supporting documents will be released and publicly available.
	
\end{itemize}

\section{Related Work}
\subsection{Surgical Video Analysis}
Most previous works treated tool presence detection and phase recognition from surgical video as two independent tasks.
The literature presenting to solve automatic tool presence detection problem in the computer assisted intervention community is relatively limited.
The early approaches were based on the low-level handcrafted features, such as the combination of shape, color and texture features (\cite{lalys2012framework}).
Recently, researchers have been dedicated to employing CNNs to learn more discriminative visual features.
Some methods proposed to recognize each kind of tools independently. 
For example, \cite{luosurgical} utilized multiple CNNs to extract the visual feature, but the performance is not satisfactory since such methods ignored the intrinsic association among different tools.
Others formulated this task as a multi-label classification problem and leveraged the underlying relationship of tools. 
\cite{wang2017deep} integrated VGG and GoogleNet to take advantage of the deep CNN model ensemble.
\cite{sahu2017addressing} paid attention to analyzing the imbalance on tool co-occurrences and exploited stratification techniques during the network training process.
\cite{choi2017surgical} developed a real-time detection CNN model based on YOLO. 
\cite{al2017surgical} proposed to leverage sequential information to detect surgical tools in cataract surgery videos, which used optical flow to fuse the multi-image during the network training.

The methods which proposed to recognize phase were mainly divided into three categories according to what types of data to be utilized, including manually annotated signal, sensor signal and the combination of them.
In addition, video data as the main sensor signals can be further separated into external OR video and endoscope video used in minimally invasive surgery. 
First, many studies leveraged various manually annotated data to recognize surgical phases.
For example, \cite{padoy2012statistical} exploited binary instrument usage signal and utilized statistical modeling based on dynamic time warping (DTW) and HMMs to analyze the data.
Forestier et al. used tool usage information, the anatomical structure, and the surgical motion which are collectively known as surgical triplets to represent frame information. Decision tree and DTW combined with a clustering algorithm were employed to process the data (\cite{forestier2013multi,forestier2015automatic}). 
These manually annotated signals can represent some typical features of phases, therefore methods based on them can achieve quite good performances.
However, this kind of signals needs additional workload which is time-consuming and tedious for surgeons.
Moreover, it cannot be obtained in real time, therefore aforementioned methods are invalid when doing a real online surgery.
In this regard, some previous works were dedicated to presenting methods which are solely based on the live sensor signal.
\cite{klank2008automatic} presented a feature extraction mechanism based on genetic programming to automatically extract visual features from surgical video.
Support vector machines were then used to classify the phases of cholecystectomy surgery from the extracted feature vectors. 
However, the average accuracy is only around 50\% in some cases due to the low level extracted features.
Considering that the methods solely based on visual features cannot reach the satisfying performance, some researchers presented approaches to leverage the live sensor signal and manually annotated signal simultaneously.
\cite{padoy2008line} proposed to combine the tool usage signals and visual cues computed from two videos, including OR video to record the surgery environment and endoscope video. A left-right HMM was constructed from these signals.
\cite{blum2010modeling} found a projection function from visual features of video frames to tool usage signals.
HMM and DTW were then utilized to model sequential dependencies.
This method can be used in test time because the tool signals are not needed anymore as long as the projection function is obtained.
However, the tool annotations are still needed in the learning process.
How to effectively and efficiently recognize surgical phase still remains an open problem.

With the advancements of high-level feature learning, many studies tended to leverage deep CNNs and RNNs to extract feature solely from online sensor signal.
Much more discriminative features contribute to the compelling performance of phase recognition and meanwhile, alleviate the challenge of high annotation workload.
\cite{dipietro2016recognizing} used RNN to model the robot kinematics and achieved an accurate phase recognition for robotic surgery.
\cite{lea2016surgical} employed temporal filters to convolve the sequential stacked spatial features extracted from CNN. DTW is then used as a classifier to recognize phase based on the spatio-temporal feature.
\cite{jin2018sv} proposed a unified framework called SV-RCNet, where CNN is utilized to extract visual features from encoscopy videos, and RNN is seamlessly integrated to model the temporal information. 
The network jointly optimized the visual representations and temporal dynamics, and achieved promising performance.
\cite{yengera2018less} presented a self-supervised pre-training approach based on remaining surgery duration prediction for addressing surgical phase recognition with less annotated data.

\subsection{Multi-task Learning}

Aforementioned methods tackled tool presence detection and phase recognition tasks separately, which cannot take the advantage of complementary information of these two tasks to benefit each other.
In addition, from some works of phase recognition (\cite{blum2010modeling,padoy2012statistical,lalys2013automatic,yu2019assessment}), it is observed that tool usage information is beneficial for recognizing phase as the input signal.
Therefore, with joint learning the phase recognition and tool presence detection, tool usage information can be indirectly used for the improvement of phase recognition through the shared features.

Recently, effectively leveraging the close correlations between multiple tasks have achieved great success in natural data analysis (\cite{mahmud2017joint,hinami2017joint,gebru2017fine,liu2017hierarchical}).
For example, \cite{mahmud2017joint} presented a multi-task network with three streams. The extracted features were concatenated for jointly inferring the activity labels and starting time.
\cite{hinami2017joint} proposed to learn a multi-task Fast R-CNN for object detection, action and attribute classification. The network shared features in earlier layers and employed a fully connected layer as a classifier in each branch.
Although achieving outstanding performance, the former method lacks shared weights to enable dynamic interaction.
While in the latter one, the branches are not well designed based on the task characteristics, and hence the intrinsic relatedness is not sufficiently exploited. 

Many studies in medical image analysis domain have also corroborated the importance of harnessing the relatedness to simultaneously improve performance of both tasks.
Multi-task learning has been demonstrating state-of-the-art results on many challenging tasks, 
such as cardiac left ventricle full quantification (\cite{xue2017full}), pulmonary nodule classification and localization regression (\cite{dou2017automated}), nuclei detection and fine-grained classification (\cite{zhou2017sfcn}), surgical instrument segmentation and localization (\cite{laina2017concurrent}), synthetic CT generation and organs-at-risk segmentation (\cite{bragmanmulti}), pancreas localization and segmentation (\cite{roth2018spatial}).
For surgical video analysis, \cite{twinanda2017endonet} recently presented a multi-task network called EndoNet to simultaneously carry out the two tasks of tool detection and phase recognition.
The network consisted of two branches that shared the early layers to extract the visual features.
Hierarchical HMM was subsequently applied to enforce the temporal constraints to refine the phase recognition results.
Although this work has achieved outstanding performance, temporal dependencies, which are crucial for phase analysis, are detached from the unified framework.
\cite{zisimopoulos2018deepphase} proposed to first train a ResNet to recognize tool presence and then combine the tool binary predictions and tool features from the last layer to train a RNN for phase recognition, which achieved promising results in cataract video analysis. 
Very recently, \cite{nakawala2019deep} present a Deep-Onto network which integrates deep models with ontology and production rules.
The method can recognize different types of surgical contexts, including phase, tool and action.
However, it is lacking in careful design of task-specific branches, which uses the multilayer perceptron for each task.
Therefore, there is still room for further investigation and improvement in terms of correlation modeling and temporal information involving.

In addition, many works have attempted to learn the relationship through a matrix space or utilize additional regularization to increase the model learning capability.
For example, \cite{augenstein2018multi} propose to leverage unlabeled or auxiliary data for better text classification. They first designed a label embedding layer to learn a relationship space between disparate labels. Based on it, a label transfer network is employed to leverage the predictions of the auxiliary tasks to estimate a label for the target task.
\cite{bachman2014learning} present an agreement regularizer to minimize variation of pseudo-ensemble models for improving sentiment analysis. They first obtain several pseudo-ensemble child models by perturbing the parent model through some noise process. Then they examine the relationship of pseudo-ensembles by the agreement regularizer and penalize the whole model.

\section{Methods}
\begin{figure*}[t]
	\centering
	\includegraphics[width=1.0\textwidth]{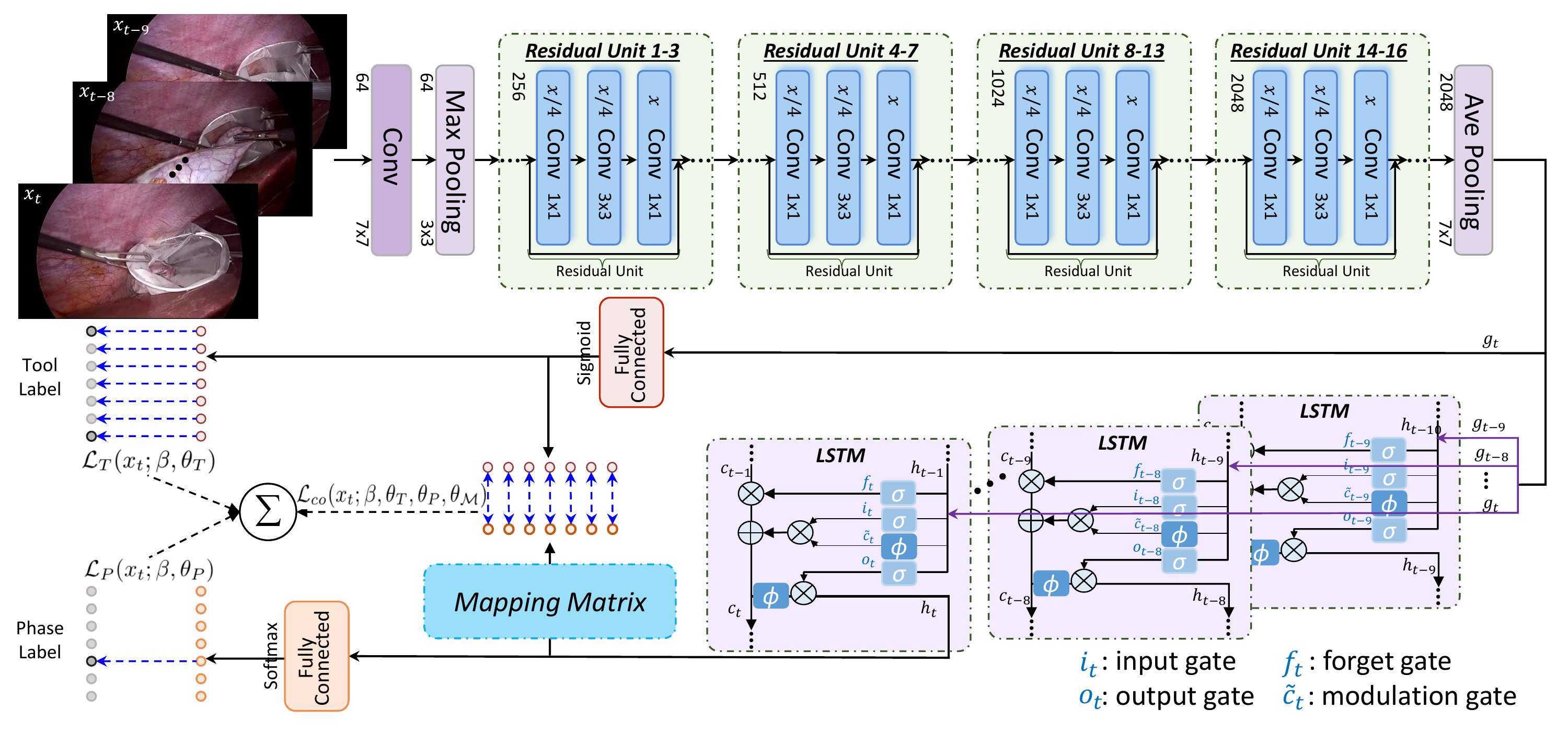}
	\caption{An overview of the proposed MTRCNet-CL for joint tool presence detection and phase recognition from surgical videos in a unified end-to-end framework. LSTM networks are illustrated by diagrams to indicate how temporal information is modeled.}
	\label{fig:overview}
\end{figure*}

Aiming to sufficiently take advantage of the natural relatedness of tool presence detection and surgical phase recognition tasks, 
we present a novel framework with two branches which share the early feature encoders and respectively hold higher layers for specific tasks.
The LSTM unit is embedded in the phase branch, which introduces the sequential dynamic into the unified framework.
In addition, we propose a correlation loss to minimize the divergence of the distributions of predicted probabilities,
thus enforcing the consistency of outputs for the two correlated tasks.
The overview of our proposed MTRCNet-CL is shown in Figure~\ref{fig:overview}.

\subsection{Multi-task Learning Network Architecture}
To meet the challenges of surgical video recognition, in the shared backbone part, we employ a 50-layer residual convolutional network to extract representative high-level features~(\cite{he2016deep}).
In each residual unit, we stack three convolutional layers, each followed by a batch normalization layer and a ReLU non-linearity layer.
After constructing the residual unit, we gradually stack 16 blocks to improve the network depth and finally form the deep residual network.
The backbone part ends with an average pooling layer and outputs a 2048-dimension feature vector.

For conducting multi-tasks, the network construction splits into two branches, respectively targeting for tool presence detection and surgical phase recognition tasks.
Considering that tool presence is defined solely based on visual information in the single frame, a fully-connected layer is directly connected to the backbone network with a sigmoid layer followed to produce predictions for the tools.
As for phase recognition which relies on temporal information, we connect the shared backbone layers with LSTM units in this branch.
There are several gates to modulate the interactions between the memory cell $c_t$ and its environment.
Hidden state $h_t$ retains the past information and supplies it to the memory cell through the gates.
The details are instantiated in the diagrams in Figure~\ref{fig:overview}. 
Different from the traditional linear models, such as HMM, our employed LSTM takes full advantage of long-term temporal information (\cite{donahue2015long}).
Moreover, to capture richer dynamics in surgical videos, we implement a distributed system enabling multiple GPUs computations, which allows us to extend the length of input sequences easily.

The tool branch (with a fully-connected CNN layer) and phase branch (with a RNN layer) are both seamlessly connected with the shared backbone convolutional layers.
Overall, we get a recurrent convolutional network to process multi-tasks on surgical videos (MTRCNet).
The entire framework is trained end-to-end supervised by both tool and phase annotations via joint learning.
By introducing temporal information in the whole training process, the framework can make use of the complementary information of visual and temporal space simultaneously.

\subsection{Objective Functions for Joint Learning of CNN and RNN}
We jointly learn the two tasks by training the framework with extracted video clips.
We denote each video clip input to the network by $\bm{x} \! = \! \{ x_{t'}, \ldots, x_{t-1}, x_t \}$, where $x_{t'}$ is the first frame and $x_t$ is the last frame in this clip sample. 
The number of frames in each video clip is represented as $N_f$. 
Meanwhile, we denote the shared layers by $U$ with weights $\beta$ and its obtained feature vector for frame $x_t$ is represented by $g_t$.
The stacked higher layers in two branches are respectively represented by $V_T$ for tool and $V_P$ for phase with weights $\theta_T$ and $\theta_P$, respectively.

We treat the tool presence detection as a multi-label classification problem, given that different categories of tools may appear in the same frame.
To this end, we utilize multi-label logistic loss to calculate the classification error for the tool branch.
Denoting the $C$ as the set of tool categories, the tool branch loss function is defined as follows:
\begin{equation} 
	\label{eq:tool}
	\mathcal{L}_{T}(x_t;\beta,\theta_T)= - \sum\limits_{c\in C} (y^T_{t,c} \log(\hat{p}_{t,c}) + (1 - y^T_{t,c}) \log(1 - \hat{p}_{t,c})), 
\end{equation}
where $y^T_{t,c} \! \in \! \{0,1\}$ is the ground truth of tool presence for frame $x_t$, which equals to 1 when the $c$-th tool presents in the $t$-th frame;
$\hat{p}_{t,c}$ represents the prediction of the $c$-th tool presenting in frame $x_t$.

In the phase recognition task, we use softmax cross-entropy function to calculate the loss of this multi-class classification task:
\begin{equation}
	\label{eq:phase}
	\mathcal{L}_{P} (x_t ;\beta, \theta_P) = -  \log \hat{p}_t^{z=y^P_t}(x_{t':t},h_{t':t-1} ),
\end{equation}
where $\hat{p}_t^z$ represents the predicted probability of frame $x_t$ belonging to the phase class $z$;
$y^P_t$ is denoted as the phase ground truth label of frame $x_t$;
$h_t$ indicates the updated hidden state calculated by LSTM with input frame $x_t$ and previous hidden state $h_{t-1}$.
With such recurrent module in the unified framework, sequential dynamics in the video are jointly learned with visual representations.

Training the entire framework in an end-to-end manner enables to simultaneously and interactively recognize the tool and phase.
With shared visual features extracted by the earlier layers and joint optimization of two branches, the learning of both tasks can benefit from each other.
Specifically, according to Eq.\ref{eq:tool} and Eq.\ref{eq:phase}, the shared weights $\beta$ in the earlier convolutional layers $U$ are optimized by both tool presence detection loss $\mathcal{L}_{T}$ and phase recognition loss $\mathcal{L}_{P}$.
The gradients derived from tool branch can flow to the layers in phase branch, and vice versa.
More importantly, the recurrent module LSTM in phase branch brings the temporal information to the unified network, which can be jointly learned with the shared convolutional module.
To this end, temporal information not only has a positive effect on the phase recognition, but also implicitly benefits the tool detection.

\subsection{Correlation Loss Modeling Relatedness between Tasks}
In surgery, surgeons are requested to perform some specified operations with corresponding tools in a specific surgical phase.
Therefore, the tool presence and surgical phase have well-defined prior correlations with clear domain knowledge.
In this regard, the classical multi-task learning network which learns the shared features at low-level layers while uses task-specific high-level predictors, such as our afore-defined MTRCNet, is suboptimal for this problem.

\begin{figure*}[t]
	\centering
	\includegraphics[width=0.85\textwidth]{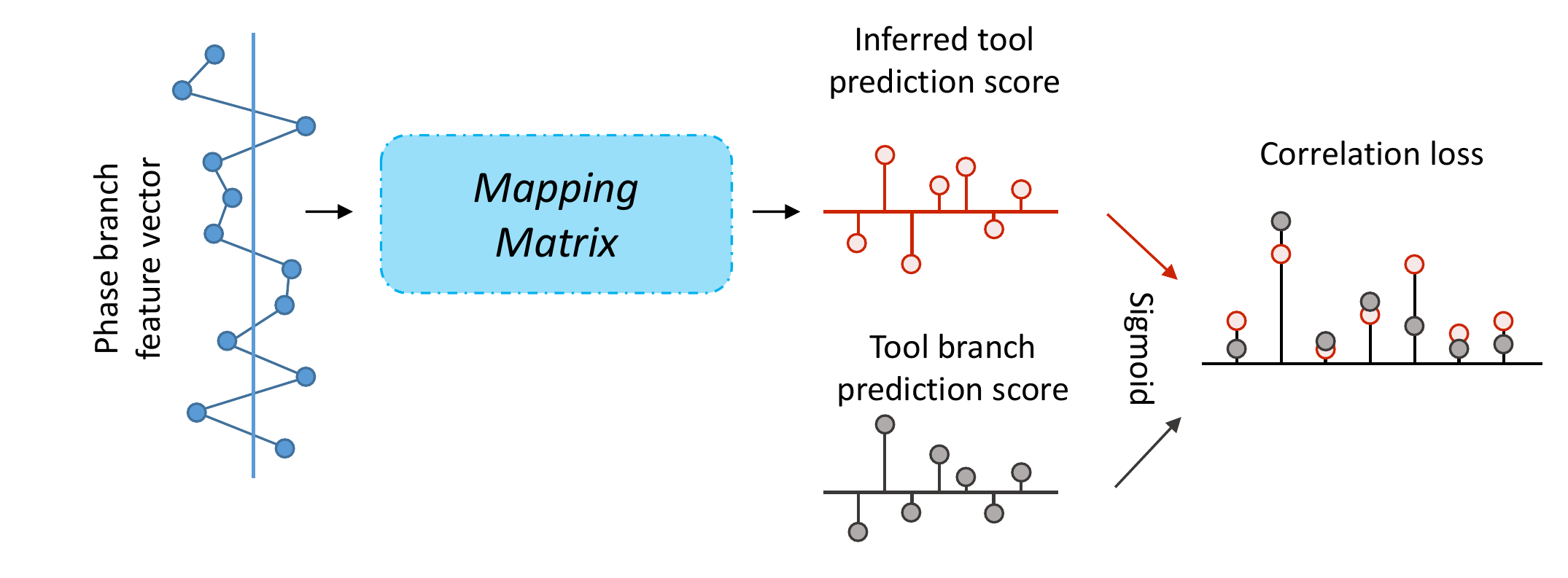}
	\caption{Illustration of the calculation process of the proposed correlation loss.}
	\label{fig:correlationloss}
\end{figure*}

Previous multi-task method (\cite{twinanda2017endonet}) for surgical video analysis improves the traditional architecture by directly concatenating tool predictions with visual features to conduct phase recognition.
Instead, we design more carefully about how to more effectively model the relatedness between the two tasks.
Specifically, we argue that we can also make reliable predictions for tool presence, by only using the features from the phase recognition branch, as there exist underlying mapping patterns between the two label spaces.
Moreover, this obtained probability distribution of tool presence inferred via the phase features, can serve as a referenceable prior to regularize the predictions of the tool branch.
With this analysis, we construct a correlation cell, i.e., a mapping matrix with $128 \times 7$ dimensions, to translate the underlying correlation between these two tasks. 
Practically, this mapping matrix is to linearly cast the high-dimensional spatial-temporal features to a compact semantic space with the meanings of surgical tools. 
Furthermore, the divergence of the probability distributions of the tool usage is minimized via a derived correlation loss,
penalizing the inconsistency between the tool branch and the inferred prior.
The Figure~\ref{fig:correlationloss} illustrates the concepts of this process.

We choose the Kullback-Leibler (KL) divergence to establish this additional regularization, with considerations on particular characteristics of our problem.
Usually, it is common to use mean square error (i.e., L2 Norm) to measure the Euclidean distance of two vectors, especially in scenarios of regression problems. However, measuring L2 Norm is inappropriate in our problem setting, as we are not enforcing the equality of absolute values of probability predictions.
We are instead expecting to measure the distance of two probability distributions.
The cross-entropy loss is also not optimal, given that the two distributions are both unfixed, which would result in failure in measuring the absolute difference if using cross-entropy loss.
Meanwhile, the Earth-Mover distance (i.e., Wasserstein-1 distance), which is widely used and brings in stability for generative adversarial networks, is not suitable for our task. 
It is good at tackling the situation where the distribution's support does not have non-negligible intersection, and when KL-divergence is just infinite.
In our setting, the two predictions are highly correlated, such that KL-divergence can be smoothly used to measure the difference between two distributions.
In these regards, we derive the correlation loss for our multi-task learning based on KL-divergence.
We regard the two distributions obtained from both branches as equally important, and therefore, we compute the KL-divergence bi-directionally.
We choose not to use the Jensen-Shannon (JS) divergence (symmetrical distance), with consideration that the distance is calculated towards the average of two distributions; such mixture has no intuitive meaning in the real-world application.
Instead, the bi-directional KL-divergence directly computes on the tool and phase distributions, which is a more straight and stable implementation.

Formally, with $r_t$ denoting the features output from LSTM in phase branch for frame $x_t$, the tool prior probability $\tilde{p}_{t,c}$ can be inferred by forwarding $r_t$ to the mapping matrix $\mathcal{M}$ with parameters $\theta_\mathcal{M}$, followed by a \textit{Sigmoid} activation:
\begin{equation}
	\label{eq:mapping}
	\tilde{p}_{t,c} = \textit{Sigmoid}(\mathcal{M}(r_t;\theta_\mathcal{M})).
\end{equation}
We denote the predicted probability distribution of tool $c$ obtained in the tool branch by $\hat{p}^\prime_{t,c} = [\hat{p}_{t,c}, 1- \hat{p}_{t,c}]$, and the prior inferred from the phase branch by $\tilde{p}^\prime_{t,c} = [\tilde{p}_{t,c}, 1- \tilde{p}_{t,c}]$.
The correlation loss of each tool is calculated bidirectionally with KL divergence as $\mathcal{D}_{KL} (\hat{p}^\prime_{t,c} \| \tilde{p}^\prime_{t,c})$, and $\mathcal{D}_{KL} (\tilde{p}^\prime_{t,c} \| \hat{p}^\prime_{t,c})$. Overall, the correlation loss is the sum of those for all categories of tools:
\begin{equation}
	\label{eq:kl}
	\centering
	\begin{gathered}
		\mathcal{L}_{co}(x_t;\beta,\theta_T,\theta_P,\theta_\mathcal{M})=\sum\limits_{c \in C} (\frac{1}{2} \mathcal{D}_{KL} (\hat{p}^\prime_{t,c} \| \tilde{p}^\prime_{t,c})  +  \frac{1}{2} \mathcal{D}_{KL} (\tilde{p}^\prime_{t,c} \| \hat{p}^\prime_{t,c}) ), \\
		\mathcal{D}_{KL}(\hat{p}^\prime_{t,c} \| \tilde{p}^\prime_{t,c})= \hat{p}_{t,c} \log\frac{\hat{p}_{t,c}}{\tilde{p}_{t,c}} + (1-\hat{p}_{t,c}) \log\frac{1-\hat{p}_{t,c}}{1-\tilde{p}_{t,c}} .\\
	\end{gathered}
\end{equation}
By enhancing the consistency between the two predictions, the weights for each task are not only optimized by corresponding ground truth, but also influenced by the information from the other related task.
In other words, the correlation loss forces the phase branch to encode tool presence information into feature vectors $r_t$, and meanwhile it constrains tool branch to take into account phase representation by enforcing the tool branch to learn from the perspective of phase.
In addition, the correlation loss provides additional regularization and supervision to improve the interactions when updating the weights $\theta_T$ and $\theta_P$ in two branches.
By encouraging the joint optimization of tool and phase branch, the correlation between the two tasks can be further captured
and modeled in the correlation cell.
The updated correlation cell provides more accurate tool prediction inference from phase branch, which forms a beneficial circulation for the entire network training.

\subsection{Overall Loss Function, Training Procedure and Implementations}
\label{imple}
We denote $N_c$ as the number of clip samples in the whole training database and each sample contains $N_f$ frames.
The parameters in the entire framework is represented by $W=\{\beta,\theta_P,\theta_T,\theta_\mathcal{M} \}$.
The overall joint loss function then can be formulated as:
\begin{equation}
	\label{eq:overall}
	\mathcal{L} = \frac{1}{N_c N_f} \sum\limits_{i=1}^{N_c} \sum\limits_{t=1}^{N_f} (\mathcal{L}_{T}(x_{t,i}) + \lambda_1\mathcal{L}_{P}(x_{t,i}) + \lambda_2\mathcal{L}_{co}(x_{t,i})  )  +\lambda_3\|W\|_2^2,
\end{equation}
where $x_{t,i}$ is the $t$-th input frame in the $i$-th video clip sample;
the first three terms represent tool detection loss, phase recognition loss and correlation loss, respectively;
the last term corresponds to the weight decay regularization.
The $\lambda_1$, $\lambda_2$ and $\lambda_3$ are hyper-parameters to balance the loss.
We employ stochastic gradient descent method to jointly update weights of the entire framework.
With shared low-level feature extracted in earlier convolutional layers and high-level constraint using correlation loss and mapping matrix,
the dynamic interaction between two branches can be facilitated during the joint training procedure.
To this end, our MTRCNet-CL can sufficiently leverage the close relatedness between the two tasks, and hence improve the performance of both tasks.

To sufficiently take advantages of relatedness between the two tasks, it is necessary to carefully design the training procedure.
In practice, we exploit a three-step strategy to train our framework.
In step-1, given that the parameter scale of backbone shared layers is much larger than that of two branches,
we initialize the weights of backbone shared layers with a pretrained model on ImageNet~(\cite{he2016deep}).
The branch-specific weights $[\theta_T,\theta_P]$ are randomly initialized with Xavier uniform initializer.
Then the MTRCNet with two branches is jointly trained with $\mathcal{L}_{T}$ and $\mathcal{L}_{P}$.
In step-2, we freeze $[\beta,\theta_T,\theta_P]$ and solely train the mapping matrix $\theta_\mathcal{M}$ from phase feature towards tool labels.
After obtaining a reliable mapping matrix which is able to construct the close relatedness of two tasks, in step-3, we jointly optimize the entire parameters of MTRCNet $[\beta,\theta_T,\theta_P]$ and the weights of mapping matrix  $\theta_\mathcal{M}$ towards Eq.~\ref{eq:overall}, i.e., the overall loss function  $\mathcal{L}$ with correlation loss added. 
Note that when the two types of annotations are unequal, we can divide the jointly training in step-1 into two independently training process of tool and phase branch with corresponding subsets, which can assist to make full use of the data annotations.
During the testing inference, the tool and phase predictions are output by the two network branches. 
The tool predictions from the mapping matrix are not utilized or averaged with ones from the tool branches, as the performances have no obvious improvement when being evaluated on the validation dataset.
As shown in the experimental results, the designed training and testing strategy delivers an outstanding performance.

In implementation, we first down-sample the original videos to enrich the temporal information in one input video clip. 
We choose to down-sample the video from 25\emph{fps} to 1\emph{fps} considering that tool presence is annotated in 1\emph{fps}.
We resize the frames from the original resolution of $1920 \! \times \! 1080$ and $854 \! \times \! 480$ into $250 \!  \times \!  250$ to dramatically save memory and reduce network parameters.
The data augmentations with $224\times224$ cropping and mirroring is performed to enlarge the training dataset.
We train the model using back-propagation with stochastic gradient descent, with the momentum of 0.9 and weighted decay of $5e\!-\!4$.
We initialize learning rate of shared convolutional layers as $5e\!-\!5$ and two sub-branch layers as $5e\!-\!4$, and are divided by a factor of 10 when the validation loss plateaus.
Our framework is implemented based on PyTorch with 4 NVIDIA Titan Xp GPUs for training. 
Such implementation of multiple GPUs enables the input length of each video clip to reach $10$ seconds and enables the batch size to reach $400$.
It takes around 8 hours for training the entire framework.
We use one GPU configuration in test inference.

\section{Experiments}
\subsection{Dataset and Evaluation Metrics}
We extensively validate the proposed MTRCNet-CL method on a large public surgical dataset, i.e., Cholec80 (\cite{Cholec}).
The dataset consists of $80$ videos recording the cholecystectomy procedures.
The videos are obtained at 25~\emph{fps} and each frame has the resolution of $1920 \times 1080$ or $854 \times 480$.
All the frames are annotated with $7$ defined phases by experienced surgeons.
Tool annotations, also consisting of 7 categories, are conducted at 1~\emph{fps} re-sampling.
The tool presence is defined based on the visual information of a tool in the single frame and annotated as a positive one if at least half of the tool tip is visible.
The detailed definition and typical appearance of phases and tools in the Cholec80 dataset are presented in Figure~\ref{fig:relation}.
Following the same procedure reported in~\cite{twinanda2017endonet}, we split the dataset Cholec80 into two subsets with equal size, with $40$ videos for training and the rest $40$ videos for testing.
All our experiments are conducted in online mode, i.e., without using future information $\{x_{t+1}, x_{t+2},...\}$ when making predictions for frame $x_t$.

To quantitatively evaluate the performance of our method, we employ the evaluation metrics utilized in~\cite{twinanda2017endonet}.
For phase recognition, we use precision (PR), recall (RE) and accuracy (AC) to validate the performance.
The PR and RE are computed in phase-wise, defined as:
\begin{equation}
	\label{eq:eva}
	\centering
	\begin{gathered}
		\mathrm{PR}=\frac{|\mathrm{GT} \cap \mathrm{P}|}{|\mathrm{P}|}, ~ \mathrm{RE}=\frac{|\mathrm{GT} \cap \mathrm{P}|}{|\mathrm{GT}|},\\
	\end{gathered}
	\small
\end{equation}
where $\mathrm{GT}$ and $\mathrm{P}$ represent the ground truth set and prediction set of one phase, respectively.
After PR and RE of each phase are calculated, we average these values over all the phases and obtain the PR and RE of the entire video.
The AC is calculated at video-level, defined as the percentage of frames correctly classified into the ground truths in the entire video.
For tool recognition, the performance is evaluated by mean average precision (mAP).
We first calculate the AP of each tool and average them over all the seven tools.
In the following result tables, we list the average and standard deviation values computed in all the test videos, to show the mean and variation among different surgical videos.

\subsection{Effectiveness of Key Components of MTRCNet-CL}

\begin{table}[t]
	\centering
	\caption{Experimental results of ablation analysis for different network components.}
	\vspace{-2mm}
	\label{tab:multi}
	\renewcommand{\arraystretch}{1.2}
	\begin{center}
		\setlength{\tabcolsep}{4.1pt}
		\resizebox{0.9\textwidth}{!}{
			\begin{tabular}{lccccc}
				\hline
				\multirow{2}{*}{Method}  & \multirow{2}{*}{Length(s)}   & \multicolumn{3}{c}{Phase}  & \multicolumn{1}{c}{Tool} \\
				\cmidrule{3-5} \cmidrule{6-6}
				& & Precision        & Recall    & Accuracy    & mAP \\ \hline
				\multirow{3}{*}{SingleNet}
				& 4 & $81.3\pm{5.9}$  & $81.9\pm{9.3}$ & $85.3\pm{6.9}$ & \multirow{3}{*}{$85.4\pm{8.3}$} \\  \cmidrule{2-5}
				& 6  & $81.3\pm{9.2}$  & $83.2\pm{6.8}$ & $85.7\pm{7.2}$ &  \\ \cmidrule{2-5}
				& 10  & $82.9\pm{5.9}$  & $84.5\pm{8.0}$ & $86.4\pm{7.3}$ &  \\ \hline
				\multirow{3}{*}{MTRCNet}
				& 4 & $82.5\pm{5.9}$  & $82.2\pm{9.0}$ & $85.9\pm{7.6}$ & $86.4\pm{7.8}$ \\  \cmidrule{2-6}
				& 6  & $82.6\pm{6.4}$  & $84.1\pm{9.9}$ & $86.7\pm{7.2}$ & $86.5\pm{7.3}$ \\ \cmidrule{2-6}
				& 10  & $85.0\pm{4.1}$  & $85.1\pm{7.1}$ & $87.3\pm{7.4}$ & $87.5\pm{7.6}$ \\ \hline
				MTRCNet-CL & 10  & $\pmb{86.9\pm{4.3}}$  & $\pmb{88.0\pm{6.9}}$ & $\pmb{89.2\pm{7.6}}$ & $\pmb{89.1\pm{7.0}}$ \\ \hline
			\end{tabular}
		}
	\end{center}
\end{table}

We conduct extensive ablation experiments to validate the effectiveness of different key components in the proposed MTRCNet-CL model.
In Table 1, we list the results of three configurations:
(1) we independently train two networks for tool presence detection and phase recognition (SingleNet in Table~\ref{tab:multi}) as the baselines of our experiments,
where we employ the same network architectures as used in the MTRCNet-CL to guarantee the comparison fairness;
(2) we train the multi-task network with two branches in an end-to-end manner, but without any correlation loss, i.e., MTRCNet which follows classical multi-task learning practice;
(3) we add the correlation loss to the multi-task learning framework to unleash the relatedness between two tasks to a large extent, i.e., our proposed MTRCNet-CL method.
\\
\\
\textbf{Benefits of Video Length.}
The length of the input video clip has the beneficial influence on the quality of the temporal features learned from the LSTM, therefore it is considered to be a key factor for accurate phase recognition.
In order to lengthen the input video and thus enhance the temporal representation capability of our model, we implement our networks in a distributed way with multiple GPUs.
We first conduct experiments using three different input lengths, i.e., $4$, $6$ and $10$ seconds, with both SingleNet and MTRCNet to validate the effectiveness of increasing the length of the input video.

In Table~\ref{tab:multi}, we can observe that the phase recognition results produced by the single phase network (SingleNet) gradually improve with the increase of the video input length.
In particular, the metric AC improves from $85.3\%$ to $86.4\%$ when the length increases from $4$ seconds to $10$ seconds,
demonstrating the importance of learning long-term temporal dependencies for phase recognition task.
The beneficial impact can also be witnessed when we use multi-task learning architecture (MTRCNet), increasing the AC of phase recognition from $85.9\%$ to $87.3\%$.
To take advantage of the temporal information from long video clips, we employ 10-second videos as inputs for the MTRCNet-CL.
\\
\\
\textbf{Effectiveness of Multi-task Learning.}
For phase recognition task, compared with the counterparts using SingleNet, our proposed MTRCNet, although with absence of correlation loss, can achieve consistent improvements in all three evaluation metrics across all different input lengths.
The PR, RE and AC of MTRCNet with 10-second video input reach $85.0\%$, $85.1\%$ and $87.3\%$, respectively.
For tool presence detection task, MTRCNets also achieve better performance compared with the independently trained network.
In addition, it is observed that the increase of input length of MTRCNets also benefits the results of tool recognition task via the multi-task learning.
The underlying reason is that, the positive effect of longer temporal dependencies for phase recognition can be transferred into the shared  spatial-temporal features in earlier layers. 
By jointly training with two branches, the more discriminative spatial-temporal features in the shared layers can benefit the tool detection task to some extent. 
More importantly, with correlation loss added to enforce the prediction consistency, the results of our MTRCNet-CL for both phase and tool tasks are further improved, peaking at $89.2\%$ AC for phase and $89.1\%$ mAP for tool.
This demonstrates that the high-level constraint assists the network to further capture the intrinsic relatedness through penalizing the difference of prediction from two branches, and then benefits the training of both branches.
We also evaluate tool predictions from the mapping matrix of our MTRCNet-CL, achieving $88.4\%$ mAP.
It indicates that through leveraging the high correlation, we can infer a reliable prior for tool from the phase features, forming a strong base for the correlation loss calculation.
We further average these two outputs for tool recognition, with $88.8\%$ mAP performance obtained, which is not as good as the results directly output from tool branch.

\begin{figure*}[t]
	\centering
	\includegraphics[width=1\textwidth]{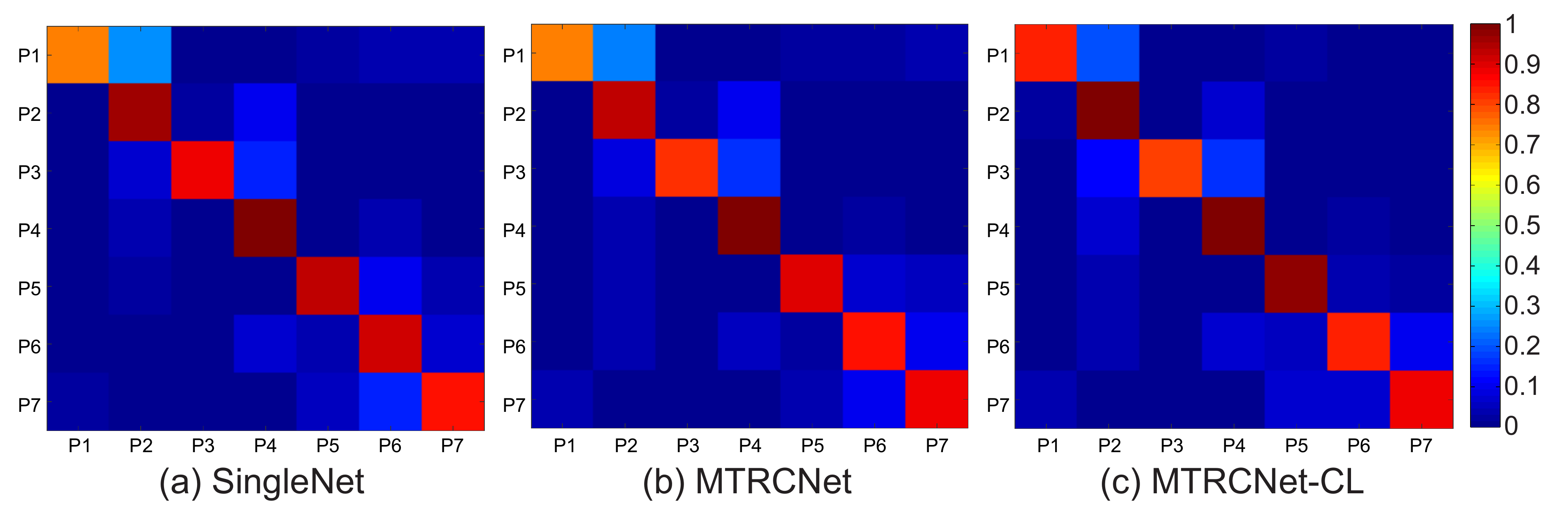}
	\caption{Confusion matrices visualized by the color brightness of three methods
		(a) SingleNet, (b) MTRCNet, and (c) MTRCNet-CL.
		In each confusion matrix, the X and Y-axis indicate predicted phase label and ground truth, respectively; 
		element (x, y) represents the empirical probability of predicting class x given that the ground truth is class y;
		the probability number on diagonal is the recall for each surgical phase.}
	\label{fig:confusion_matrix}
\end{figure*}

In order to more comprehensively analyze effectiveness of the proposed multi-task learning scheme, we first visualize the confusion matrices of phase recognition results which can show the details in phase level.
Specifically, the confusion matrices of three methods, i.e., SingleNet, MTRCNet and MTRCNet-CL with 10-second video input, are illustrated in Figure~\ref{fig:confusion_matrix}.
We omit the detailed phase names and do further abbreviation, e.g. Phase 1 to P1, to increase the concision and aesthetics of table.
We can observe that from (a) to (c), the probability percentages on diagonals (recall) tend to increase with misclassification gradually decreasing.
Particularly, it is clearly shown that the condition of incorrectly recognizing P1 into P2, P6 or P7 is consistently alleviated by using our  multi-task learning scheme. 
The same situation can also be witnessed in P5 recognition process from Figure~\ref{fig:confusion_matrix}.
These observations demonstrate that joint training with tool annotations is of great benefit to increasing recognition performance of some phases.
Apart from low-level features shared in the early convolutional layers, our high-level correlating mechanism is capable of reinforcing the interaction between two branches and can further enhance the leverage of both annotations and improve the performance.

\begin{figure*}[t]
	\centering
	\includegraphics[width=1\textwidth]{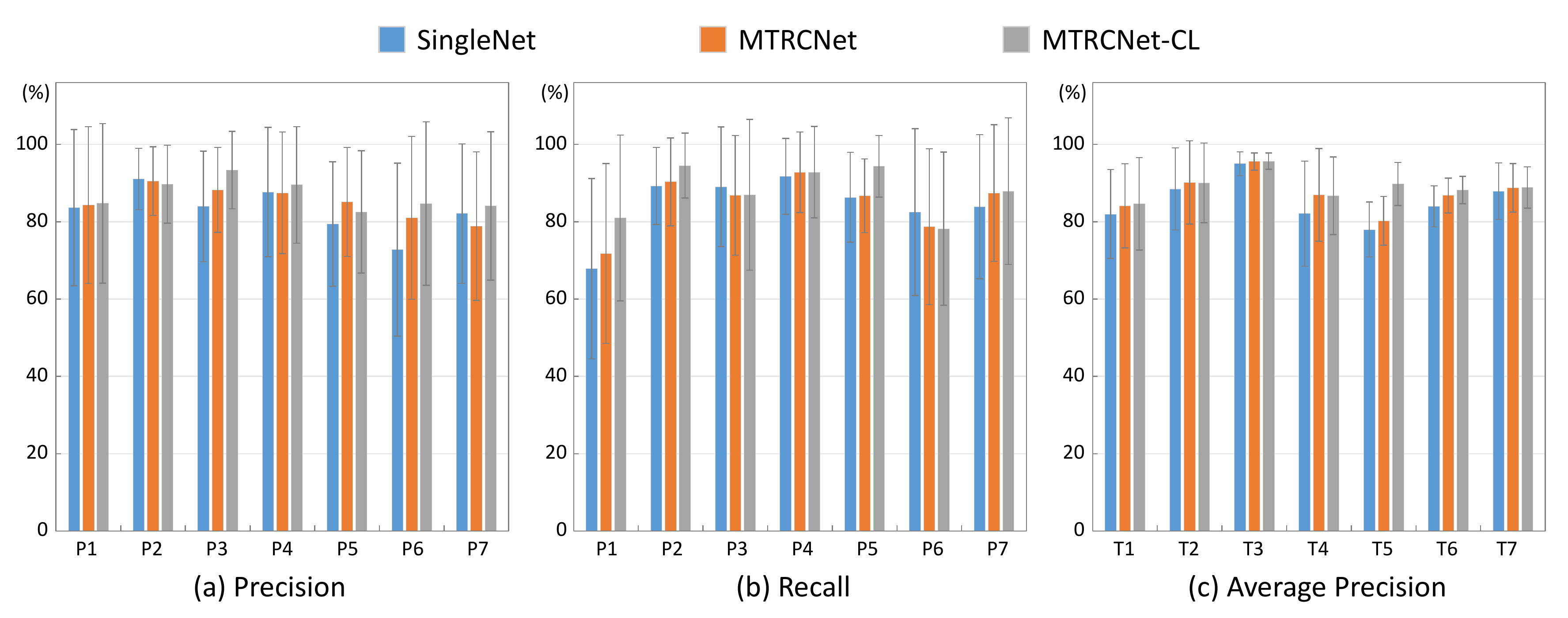}
	\caption{The bar chart results of (a) Precision and (b) Recall in phase-level and (c) Average Precision in tool-level of three methods:
		SingleNet, MTRCNet, and MTRCNet-CL. The standard deviations are shown through the error bars in each chart.}
	\label{fig:bar}
\end{figure*}

We further draw bar charts (see Figure~\ref{fig:bar}) in order to detailedly illustrate the results of PR and RE in each phase-level and AP in each tool-level.
Tool names are also abbreviated, for example, from Tool 1: Grasper to T1.
It is observed that compared with other two schemes, the MTRCNet-CL improves the PR performances in the most phases, especially in P3. 
Similarly, RE performances in P1 and P5 have especially significant increase by using MTRCNet-CL.
For AP, the MTRCNet-CL dominates other two schemes across all the seven tools.

\subsection{Detailed Ablation Study of Mapping Matrix and Correlation Loss}
The learned mapping matrix infers the tool predictions from the features in phase branch, and plays the key role on the calculation of the correlation loss.
Therefore, the configurations of mapping matrix, such as training strategy, position and mapping direction, have an important effect on the effectiveness of the correlation loss and indirectly determines the final performance of our MTRCNet-CL.
In this regard, we establish several schemes to obtain a comprehensive insight on how the mapping matrix influences the interaction of the two tasks.
All experiments have the same network architectures and data augmentation strategies for fair comparison.
The experimental results are listed in Table~\ref{tab:ablation_mapping}.
\\
\begin{table}[t]
	\centering
	\caption{Experimental results of ablation analysis for mapping matrix and correlation loss.}
	\vspace{-4mm}
	\label{tab:ablation_mapping}
	\renewcommand{\arraystretch}{1.2}
	\begin{center}
		\setlength{\tabcolsep}{3pt}
		\resizebox{0.9\textwidth}{!}{
			\begin{tabular}{l|ccc|c}
				\hline
				\multirow{2}{*}{Different Settings}    & \multicolumn{3}{c}{Phase}  & \multicolumn{1}{|c}{Tool} \\
				\cmidrule{2-4} \cmidrule{5-5}
				& Precision        & Recall    & Accuracy    & mAP \\ \hline
				Training strategy 1 (TS1)    & $80.1\pm{7.1}$  & $79.2\pm{10.4}$ & $83.4\pm{8.5}$ & $85.1\pm{8.0}$ \\ 
				Training strategy 2 (TS2)    & $85.5\pm{4.6}$  & $85.6\pm{7.6}$ & $87.8\pm{7.1}$ & $88.1\pm{7.2}$ \\ \hline
				Mapping in label space & $83.5\pm{5.8}$  & $84.5\pm{8.1}$ & $87.1\pm{7.3}$ & $86.8\pm{7.1}$  \\ \hline
				Mutual mapping matrix & $84.7\pm{6.2}$  & $85.5\pm{8.9}$ & $87.4\pm{7.9}$ & $87.1\pm{7.7}$ \\ \hline
				Ours (MTRCNet-CL)  & $\pmb{86.9\pm{4.3}}$  & $\pmb{88.0\pm{6.9}}$ & $\pmb{89.2\pm{7.6}}$ & $\pmb{89.1\pm{7.0}}$ \\ \hline
			\end{tabular}
		}
	\end{center}
\end{table}
\\
\textbf{Different Training Strategies.}
We first investigate the influence of training strategies with three schemes:
(1) jointly train the two branches with mapping matrix (TS1);
(2) train the two branches firstly, next freeze two branches and train mapping matrix, and finally freeze the mapping matrix and continue to train two branches (TS2);
(3) train two branches firstly, then freeze two branches and train mapping matrix, and finally jointly train the entire network, i.e. our MTRCNet-CL (see Section~\ref{imple} for more details).
It is observed from Table~\ref{tab:ablation_mapping} that
compared with TS1, the training scheme TS2 and ours both significantly improve the AC score for phase recognition and mAP score for tool presence detection.
The difference between these schemes is that the latter two have a separate mapping matrix pre-training step, 
which provides a relatively better initialization before joint training. 
This observation validates that balancing the learning difficulties of different components in the network to a comparable level
helps to sufficiently leverage the benefit of correlation loss.
In addition, MTRCNet-CL achieves marginally better performance over training strategy TS2.
The underlying reason is that MTRCNet-CL unfreezes mapping matrix in the final stage and allows a joint training process between the mapping matrix and two branches.
Mapping matrix then can be optimized by the loss from two tasks through the back-propagation procedure.
In the three-step training strategy of our MTRCNet-CL, the performances for tool recognition reach $87.3\%$ in step-1 (from tool branch), $86.9\%$ in step-2 (from mapping matrix), and increase to $89.1\%$ in the final step (from tool branch).
Such results demonstrate that with effectively leveraging the high correlation between two tasks, a reliable prior can be obtained based on the phase feature.
In addition, the designed correlation loss can further encourage the interaction between two branches, benefiting the network to model more powerful spatio-temporal feature.

\begin{figure}[t]
	\color{blue}
	\centering
	\includegraphics[width=1\textwidth]{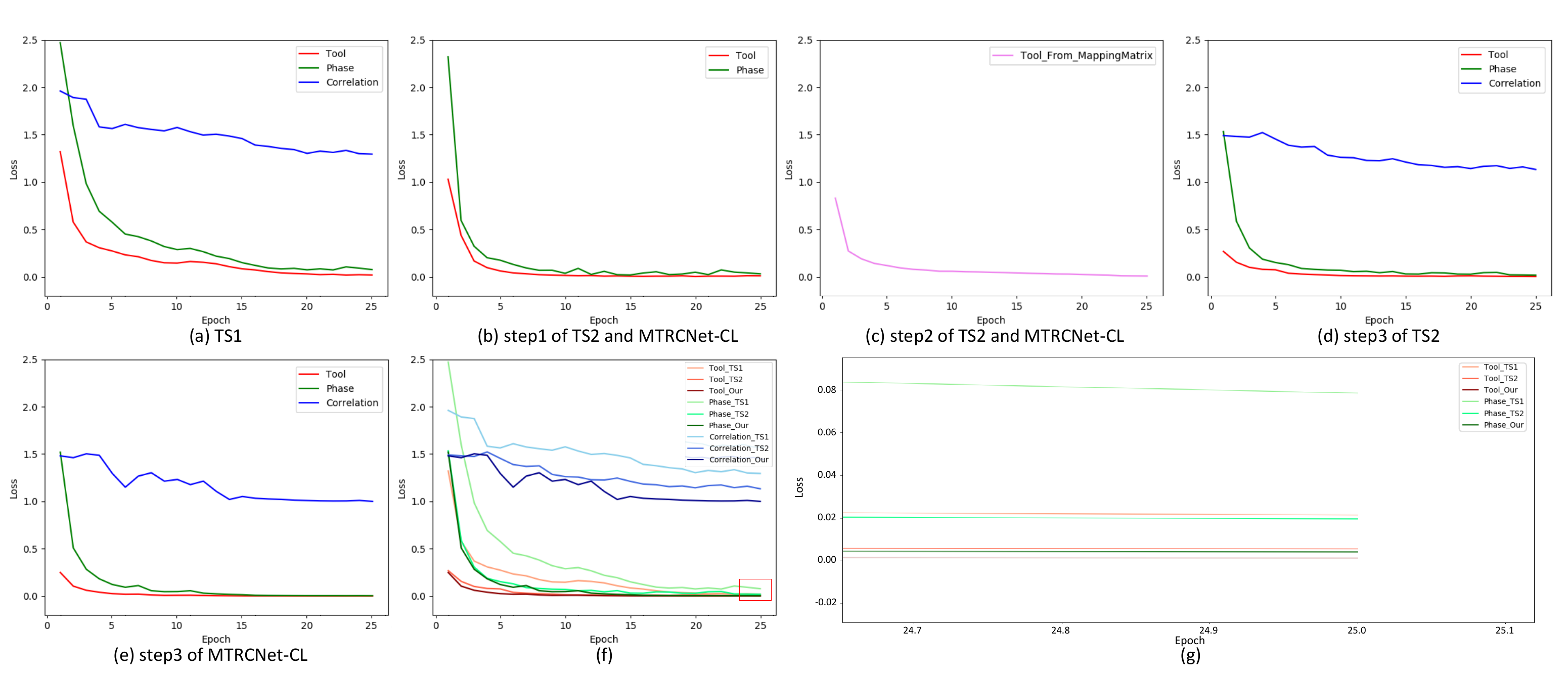}
	\caption{
		Visualization of loss curves from different steps of (a) TS1, (b c d) TS2, and (b c e) MTRCNet-CL. The loss curves of each training strategy in the last step are shown together in (f) for the clearer comparison. (g) is presented to show a closer look at the end of training process (red region of (f)). }
	\label{fig:loss}
\end{figure}
\begin{figure}[t]
	\color{blue}
	\centering
	\includegraphics[width=1\textwidth]{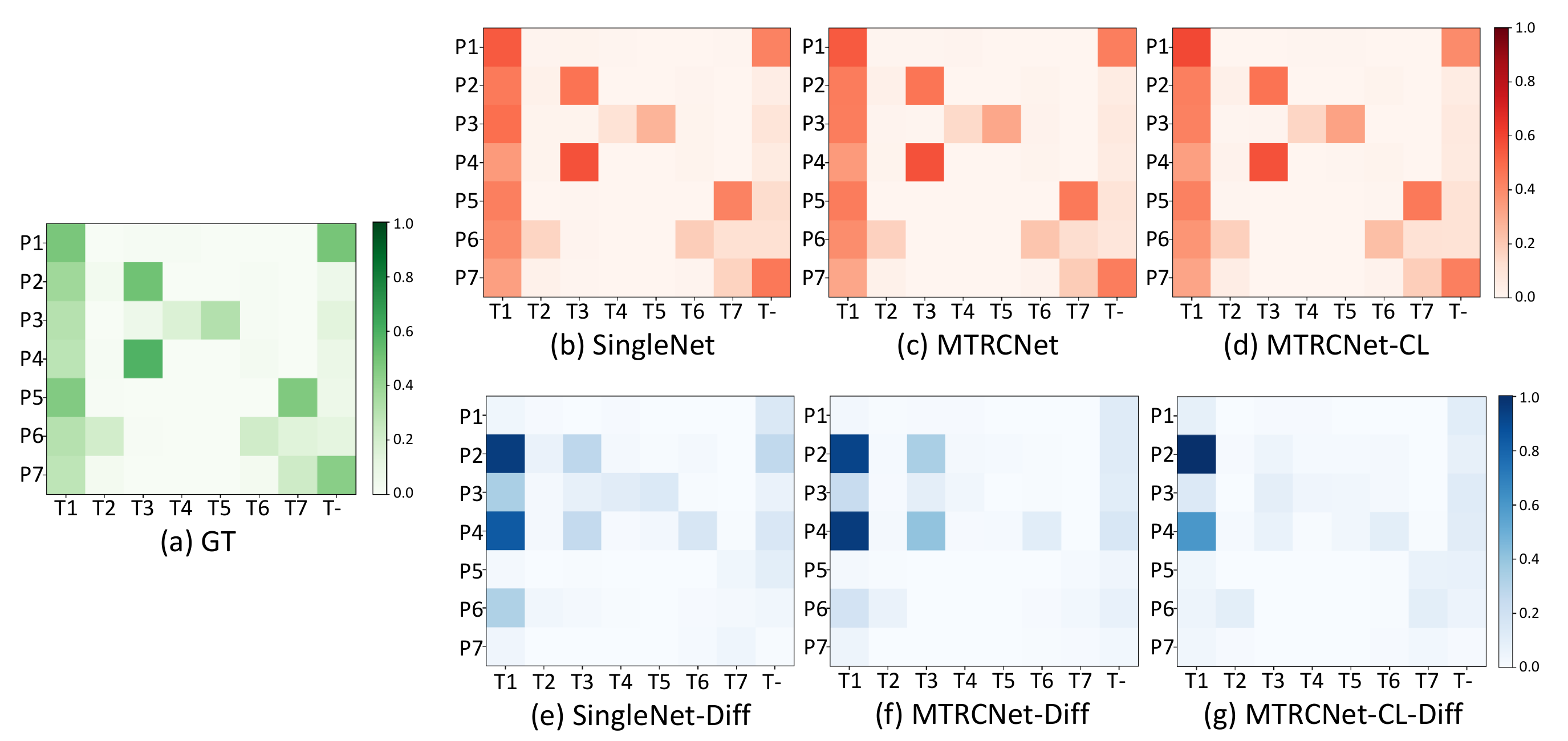}
	\caption{
		Visualization of correlation between tool and phase in Cholec80 dataset ((a) GT), and constructed correlations by our methods (b) SingleNet (c) MTRCNet and (d) MTRCNet-CL; in each correlation, element (x, y) represents the empirical probability of tool x presenting in phase y; the sum of probability numbers in each row equals to one. 
		The discrepancies between counted numbers in Cholec80 dataset and those predicted by SingleNet, MTRCNet and MTRCNet-CL are illustrated in (e), (f) and (g), respectively; the total number of tool h appearing in phase k in Cholec80 minus the total number in predictions is shown in element (h, k) of each matrix; these difference values are normalized to the range from zero to one. 
		'T-' denotes the situation that no tool appears in the frame. }
	\label{fig:correlation}
\end{figure}

We further carefully study the training processes of the three training strategies and visualize the loss plots in Figure~\ref{fig:loss}.
From (a) and (b), we observe that compared with step-1 of TS2 and MTRCNet-CL, both tool and phase losses of TS1 decrease slower when we attempt to jointly train the whole network with the mapping matrix at the beginning, which further leads to the higher loss at the end of the training process (see (f) and (g)). 
It is observed that from (d) and (e), the correlation losses consistently decrease with the tool and phase losses going down, demonstrating the beneficial complement between the learning of tool, phase branches and mapping matrix. 
In addition, our MTRCNet-CL with the designed multi-step training strategy converges much faster than using TS2, and achieves lower values of all the three losses (see (f) and (g) for the clearer comparison). It verifies the effectiveness and importance of our strategy design.
\\
\\
\textbf{Mapping in Label Space.}
We then set up the mapping matrix in label space, i.e., mapping the predicted 7-bit vector of phase recognition (from phase branch) to tool prediction.
In this regard, the ground truth of tool and phase correlation can be utilized to initialize the mapping matrix.
However, in Table~\ref{tab:ablation_mapping}, we can see that MTRCNet-CL still achieves much better performance than the label-space mapping, increasing PR and RE around $3\%$.
The difference between these two settings is that MTRCNet-CL is trained to learn a phase-feature to tool-label mapping matrix, while the other is to learn a phase-label to tool-label mapping matrix.
The latter utilized space is too compact and the learned mapping matrix is too sparse, since only one or two tools appear in each phase and the probabilities of the absent tools are around zero (see the ground truth correlation (a) in Figure~\ref{fig:correlation}).
Instead, the learned matrix of MTRCNet-CL can leverage richer information with more details in semantic feature level, and therefore can unleash the effectiveness of correlation loss.
\\
\\
\textbf{Using Mutual Mapping Matrix.}
We practically explore whether also adding a mapping matrix from tool branch to phase branch, i.e., using mutual mapping matrice between two branches, is useful.
Table~2 shows that our MTRCNet-CL consistently surpasses the scheme of mutual mapping, improving all the evaluation matrices around $2\%$.
The difference between the two schemes is the network additionally uses the mapping matrix from tool branch to phase predictions or not.
Since one tool may appear in different phases, for example, T1 has the high probabilities of appearance in all the phases, see Figure~\ref{fig:correlation} (a),
the mapping from the tool branch fails to provide an ideal prior and may even cause confusion in the joint learning.
In this regard, the configuration of our MTRCNet-CL is practically optimal to leverage the correlation loss.
Adding the extra mapping matrix from the tool branch to phase branch is practically unnecessary.
\\
\\
\textbf{Visualization of Learned Correlations of Two Tasks.}
To intuitively show the correlation existing in two tasks and provide the insight of what correlations the networks learn,
we compute the correlations between phases and tools and visualize them.
The Figure~\ref{fig:correlation} (a-d) visualizes the phase-tool correlation matrix for (a) ground truth (b)SingleNet predictions (c) MTRCNet predictions (d) MTRCNet-CL predictions through calculating the co-occurrence between the tool and phase.
The green and red color brightnesses indicating the strength of correlation for a particular phase-tool pair.
For example, we easily observe that there exist high correlations in the cholecystectomy surgical procedure, demonstrating our idea of leveraging the relatedness to improve performance has fundamental support.
With multi-task learning, both MTRCNet and MTRCNet-CL are able to reconstruct the correlations quite well, with consistency to the patterns in the ground truth.
To further clearly demonstrate the effectiveness of using correlation loss to capture the relatedness between tool and phase,
we visualize the differences between the ground truth and the predictions in Figure~\ref{fig:correlation} (e-g), with the blue color brightness indicating strength.
Note that the lower value with lighter color indicates a higher performance.
We can find that there exist largest differences between the correlations from two separately trained SingleNets and the ground truth.
The visualized difference map is noisy and scattered with large values, e.g. T1 in P6, T4 and T5 in P3.
For two multi-task learning networks, it is observed that for those close correlations with high probabilities, such as T3 in P2 and T1 in P4 (see the ground-truth correlation in (a)), MTRCNet-CL with explicitly defined correlation loss can penalize more incorrect predictions than solely using multi-branch network (MTRCNet).
This observation clearly presents the efficacy of correlation loss, with insight to retain such correlation in predictions for related tasks.

\subsection{Comparison with State-of-the-art Methods}

We compare the performance of our MTRCNet-CL with several well-known or the state-of-the-art approaches; most of the results on the same dataset are reported in~\cite{twinanda2017endonet}.
As for tool presence detection, we compare the results of our method with three approaches.
The first one is deformable part model (DPM). This method employs three components to model each tool and uses HOG features to represent the images.
The second one is 8-layer CNN (ToolNet) which is trained in a single-task way to solely perform the tool presence detection task.
The third one is a 9-layer multi-task network which leverages both tool and phase annotations (EndoNet).
This method is regarded as the state-of-the-art method for tool presence detection task in literature.

As for phase recognition, the first four comparison methods input different visual features followed by hierarchical HMM to refine the results.
They consist of 1) binary tool usage information generated from the manual annotations;
2) bag-of-word handcrafted features followed by canonical correlation analysis (CCA);
3) features extracted by 9-layer CNN which solely utilizes the phase annotations (PhaseNet); and
4) features extracted by EndoNet.
In addition, \cite{jin2018sv} proposes to seamlessly integrate CNN and LSTM to jointly learn spatial and temporal feature (SV-RCNet).
\cite{twinanda2017vision} presents to replace HHMM by LSTM to enforce the sequential constraints on the visual feature from EndoNet (EndoNet+LSTM), which achieves the state-of-the-art performance on phase recognition.
The comparison results of tool detection and phase recognition are shown in Table~\ref{tab:tool} and Table~\ref{tab:phase}, respectively.
We omit the standard deviation in these tables as not all referenced papers reported that.
F1 scores are calculated to provide the overall results of phase recognition task for better comparison.

\begin{table}[t]
	\centering
	\caption{Average precisions for recognizing the seven tools (rows) using different approaches (2nd to 5th columns).}
	\vspace{-2mm}
	\label{tab:tool}
	\resizebox{0.7\textwidth}{!}{
		\begin{tabular}{lcccc}
			\toprule
			Tool    ~~  & DPM   ~  & ToolNet    & $\text{EndoNet}^{\ast}$    & $\text{Ours}^{\ast}$ \\
			\midrule
			Grasper        & $82.3$  & $84.7$  & $\pmb{84.8}$ & $84.7$\\
			Bipolar        & $60.6$  & $85.9$  & $86.9$ & $\pmb{90.1}$\\
			Hook           & $93.4$  & $95.5$  & $\pmb{95.6}$ & $\pmb{95.6}$\\
			Scissors       & $23.4$  & $60.9$  & $58.6$ & $\pmb{86.7}$\\
			Clipper        & $68.4$  & $79.8$  & $80.1$ & $\pmb{89.8}$\\
			Irrigator      & $40.5$  & $73.0$  & $74.4$ & $\pmb{88.2}$\\
			Specimen bag   & $40.0$  & $86.3$  & $86.8$ & $\pmb{88.9}$\\ \hline
			Mean           & $58.4$  & $80.9$  & $81.0$ & $\pmb{89.1}$\\
			\bottomrule
		\end{tabular}}
		\\
		\vspace{0.5mm}
		\scriptsize Note: the $\ast$ means the methods with multi-task learning.	~~~~~~~~			
	\end{table}
	
	\begin{table}[t]
		\centering
		\caption{Phase recognition results using different approaches (rows).}
		\vspace{-3mm}
		\label{tab:phase}
		\resizebox{0.8\textwidth}{!}{
			\begin{tabular}{lcccc}
				\toprule
				Methods                   & Accuracy    & Precision    & Recall    & F1 Score \\
				\midrule
				Binary Tool               & $47.5$  & $54.4$  & $60.2$ & $57.2$  \\
				Handcrafted+CCA           & $38.2$  & $39.4$  & $41.5$ & $40.4$  \\
				PhaseNet                  & $78.8$  & $71.3$  & $76.6$  & $73.8$ \\
				$\text{EndoNet}^{\ast}$                  & $81.7$  & $73.7$  & $79.6$ & $76.5$  \\
				SV-RCNet                  & $85.3$  & $80.7$  & $83.5$ & $82.1$  \\
				$\text{EndoNet+LSTM}^{\ast}$              & $88.6$  & $84.4$  & $84.7$  & $84.5$ \\
				$\text{Ours (MTRCNet-CL)}^{\ast}$        & $\pmb{89.2}$  & $\pmb{86.9}$  & $\pmb{88.0}$ & $\pmb{87.4}$ \\
				\bottomrule
			\end{tabular}}
			\\
			\vspace{0.5mm}
			\scriptsize Note: the $\ast$ means the methods with multi-task learning. ~~~~~~~~~~~~~~~~~~~~							
		\end{table}

		In these two tables, we can find that all the CNN-based methods, including our MTRCNet-CL, achieve much higher performance than those approaches based on the handcrafted features, demonstrating that deep CNNs can extract more discriminative representations.
		Compared with two independently trained networks, i.e. ToolNet and PhaseNet, our method achieves striking improvement in both tasks, demonstrating that multi-task learning strategy is beneficial for both tool presence detection and phase recognition tasks of surgical video.
		Our MTRCNet-CL also outperforms another multi-task based method, i.e. EndoNet by a large margin.
		These comparison results verify that with low-level spatial-temporal feature sharing by CNN and RNN modules, and high-level constraint by correlation loss, 
		our MTRCNet-CL can sufficiently facilitate the interaction of two branches and therefore catch the close relatedness of two tasks.
		Moreover, our approach achieves better results than the two state-of-the-art methods, in particular, boosting the tool presence detection results from $81.0\%$ to $89.1\%$ and F1 score of phase recognition from $84.5\%$ to $87.4\%$, which corroborates the effectiveness of recurrent convolutional joint leaning and correlation loss.
		The detail results for all the seven tools are also reported in Table~\ref{tab:tool}; our approach achieves superior performances over other methods in most tool categories, especially for T4 Scissors (improving AP over $25\%$ compared with the state-of-the-art method).

		\begin{figure*}[t]
			\centering
			\includegraphics[width=1\textwidth]{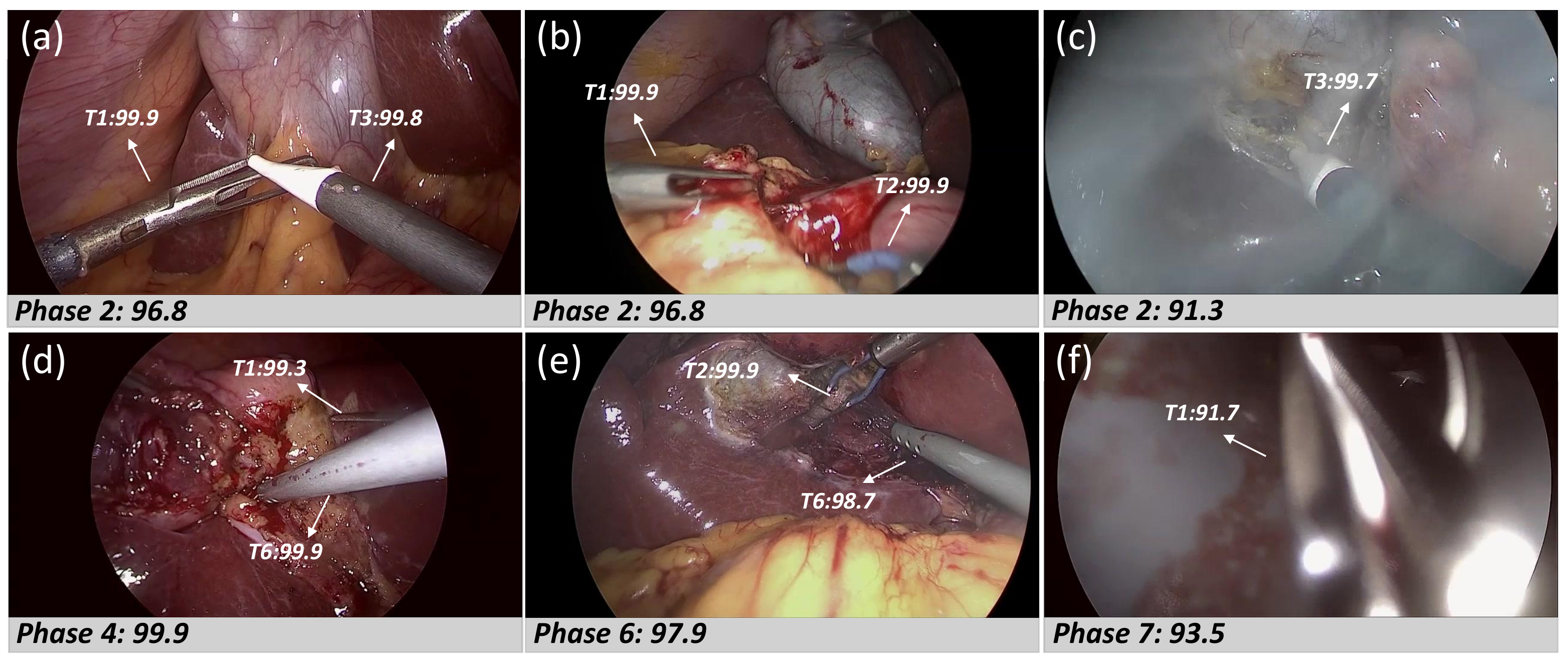}
			\caption{Typical results of the proposed MTRCNet-CL for tool presence detection and phase recognition. For tool, the labels and the probability predictions towards ground truths are indicated through white arrows; and for phase, they are present below each frame.}
			\label{fig:result}
		\end{figure*}
		
\subsection{Qualitative Results}
We present results of tool and phase recognitions in some challenging cases to illustrate the effectiveness of the proposed method, as shown in Figure~\ref{fig:result}.
Although the partial appearances and overlap of multiple tools increase the recognition difficulty (case (a), (b), (d) and (e)), 
our method can achieve rather high prediction confidence towards ground truth.
For example, T2 Bipolar in case (b) and T1 Grasper in case (d) are easy to be ignored.
However, our MTCNet-CL can witness them through the discriminative spatial-temporal feature and constructed correlation in the learned mapping matrix.
Our method is robust and able to distinguish the right phases though there exists a high intra-class variance (case (a), (b) and (c)).
Our method can identify the tools which are not normally present in some phases.
For example, T2 Bipolar is hardly utilized in Phase2 (case (b)) while T6 Irrigator is rare to appear in Phase4 (case (d)).
In other words, although the correlation between tools and phases may be unstable and complicated, our method is capable of addressing the obstacles brought from that.
In addition, our network can reduce the effect of the blur scene and noise (case (c) and (f)).

\begin{figure*}[t]
	\color{blue}
	\centering
	\includegraphics[width=1\textwidth]{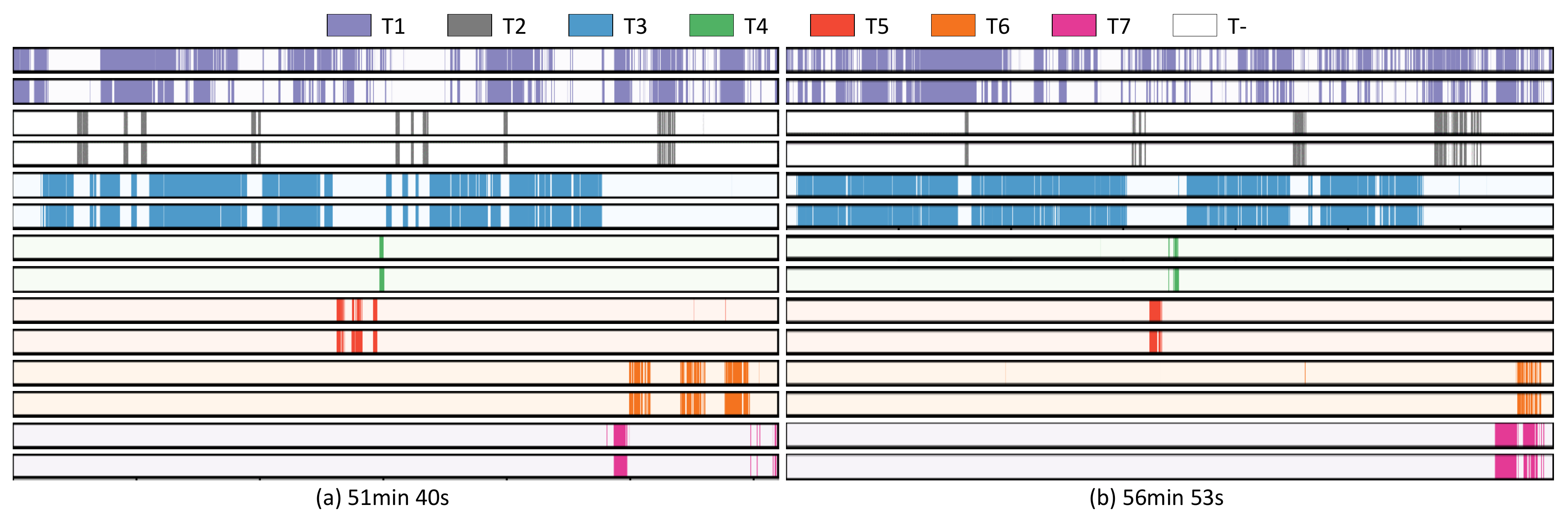}
	\caption{Color-coded ribbon visualization of tool predictions from MTRCNet-CL (above) and ground truth (bottom) in two complete surgical videos. In each case, we show the seven tool presence with different colors and no tool appearance with blank. The horizontal axes indicate the time progression of different surgical procedure, which are scaled to the same length for better visualization.}
	\label{fig:result_tool}
\end{figure*}

\begin{figure*}[t]
	\color{blue}
	\centering
	\includegraphics[width=1\textwidth]{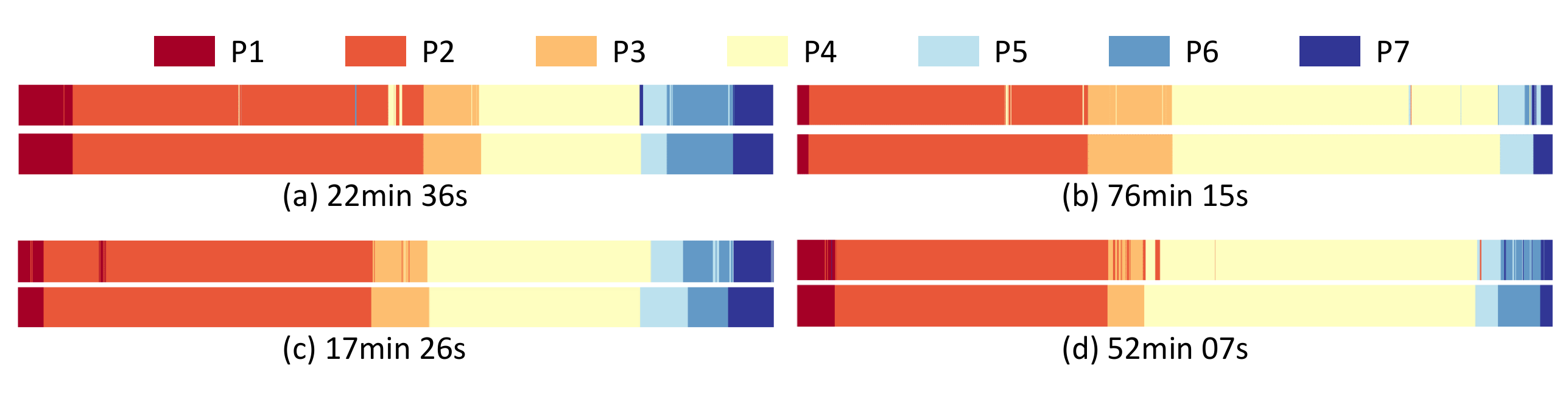}
	\caption{Color-coded ribbon illustration of phase during four complete surgical videos, whose horizontal axis represents the time progression. In each case, we present the recognition results from our MTRCNet-CL (above) and the ground truth (bottom).
		We scale the temporal axes for better visualization as the different duration in these cases.}
	\label{fig:result_phase}
\end{figure*}

We further illustrate tool and phase recognition results of several complete surgical videos in Figure~\ref{fig:result_tool} and Figure~\ref{fig:result_phase}, respectively.  
From Figure~\ref{fig:result_tool}, we observe that the presence of tool is fitful and inconsistent under the camera, even within several adjacent frames. 
This phenomenon may be caused by the quite rapid operation action and the unstable surgical camera.
Even so, our method can precisely detect different tool presences during the whole surgical procedures, demonstrating the efficacy of our multi-task learning strategy.
It can be clearly observed from Figure~\ref{fig:result_phase} that even without any post-processing method, our MTRCNet-CL can produce the smooth phase predictions with the jointly learned spatio-temporal features.
Moreover, we find that our method can accurately identify the phase transition frames, which plays a very valuable role for many computer-assisted and robotic surgery to automatically adjust the configurations and parameters to go into next phase.

\section{Discussion}

Tool presence detection and phase recognition have become a key component when developing the context-aware systems for surgical process monitoring and surgeon scheduling.
In this paper, we present a multi-task network supervised by both annotations to simultaneously address two tasks.
With the convolutional backbone shared in the earlier layers, CNN in the tool branch and RNN in the phase branch can be jointly optimized.
The entire framework is optimized in an end-to-end manner which introduces the temporal information in the whole training process.
Extensive experiments have demonstrated effectiveness of this carefully architectural design for both tasks.
More importantly, we propose a new correlation loss to provide the additional supervision by learning a mapping matrix. 
This mechanism can reinforce the interaction between two associated tasks and further enhance the high relevancy learning.
The performance improvement in the experimental results demonstrates the advantage of this additional penalization for capturing the task correlation.
A set of tailored training schemes of mapping matrix is designed to yield the maximum efficacy of the correlation loss for tackling both tool and phase tasks.
To this end, our work not only can encourage researchers to simultaneously address the associated tasks in surgical videos,
but also can inspire them to develop some strategies to facilitate the leverage of correlation for the analysis of surgical videos, as well as other interconnected multiple tasks.

According to our multi-step training strategy, in step-1, though there exist the shared earlier layers, the phase branch is jointly trained with tool branch towards the phase annotation. In step-2, the branch parameters are frozen and only the mapping matrix is trained from the modeled phase feature towards tool labels. Based on these two steps, the pathway from the phase branch focuses on the phase recognition tasks and models the spatio-temporal feature to increase the phase distinctiveness. Through the inherent high correlation learned in the mapping layer, the tool presence can be roughly inferred and derived from the modeled phase feature. Therefore, we regard the output of the mapping matrix from phase branch as an additional prior when training the whole network in step-3. During the step-3, we set hyper-parameters of phase loss and correlation loss as 1 and 0.5 respectively, to balance the loss and make the phase branch still bias to the phase recognition task. In this regard, the probability distributions obtained by the mapping matrix diverged from phase branch can still serve as a reference to tool recognition task.
In addition, the additional correlation loss have two more effects. Firstly, the two outputs to calculate correlation loss are from two possible pathways to produce tool predictions. Although one is mainly based on the tool feature and the other is mainly based on phase feature, the correlation loss between them indeed can play the ensemble role to enforce the similarity, which enhances the interaction between the two branches. Secondly, in step-3, LSTM weights can also be trained for tool recognition, which introduces some beneficial temporal information for the tool task in the form of soft label through the mapping matrix.

Although temporal information plays an essential role in the video analysis tasks,
LSTM unit is not employed in tool branch based on our careful consideration on the task definition and setting.
In the surgical video analysis community, the clinical physicians define the tool presence solely based on one frame scene without looking at adjacent frames (\cite{twinanda2017endonet}).
It is annotated as a positive one only if at least half of the tool tip is visible. 
In other words, when a tool is partially obscured in a single frame, once the visible part is less than a half, although the tool present in the surrounding frames, it will be regarded as a negative one.
In addition, the actions (e.g., hooking) in surgical videos are very rapid. Under the camera, the presence of tool is fitful and inconsistent, even within several adjacent frames. Hence, using LSTM for tool detection maybe not help improve the performance too much. 
We have conducted the preliminary experiments on this configuration with no obvious improvement shown in final performance, and even encountering a more difficult training process and suffering from longer training time.
Therefore, we choose to simplify the network architecture for computational resource saving and easier network training.

Consistency enhancement can dramatically increase the performance of phase recognition based on the prior knowledge of surgical operations.
Therefore, some works proposed to use some simple yet effective post-processing strategies to boost the final results, such as averaging smoothing in \cite{cadene2016m2cai} and prior knowledge inference, called PKI, in our previous work (\cite{jin2018sv}).
In this work, all the results are purely predicted by our end-to-end network (MTRCNet-CL) without using any post-processing strategies.
Nevertheless, we have also performed the experiment to investigate how PKI affects the results of our method, with PR, RE and AC peaking at $91.6\%$, $90.1$ and $93.3\%$, surpassing results of $90.6\%$ PR, $86.2\%$ RE and $92.4\%$ AC in \cite{jin2018sv}.

One main concern in deep learning (a data-driven methodology) is the lack of available data, especially for the surgical video.
The multi-task learning of tool and phase recognition requires the simultaneous annotations for both tasks on the same dataset, which restricts the development to a certain extent.
Fortunately, most works regard it as a worthy trade-off: some label the two tasks with utilizing binary tool usage to address phase recognition task (\cite{padoy2012statistical,yu2019assessment}); others are dedicated to establishing more advanced multi-task strategies (\cite{zisimopoulos2018deepphase,nakawala2019deep}).
In addition, more relevant datasets begin to be released for public usage, which alleviate the annotation problem to a great extent (\cite{nakawala2019deep,miccai2018}).
We choose a large-scale and well-organized dataset (Cholec80) to validate our MTRCNet-CL. The outstanding results on this typical dataset demonstrate the effectiveness of our method for surgical tool and phase recognitions.
More importantly, compared with other aforementioned multi-task methods, our network has the capability to be extended for semi-supervised learning with less annotations.
The correlation loss can be utilized as an unsupervised loss for the unlabeled data. We will explore this promising direction in the future.

Our proposed method can recognize tool and phase at a quick speed (around 0.3s per frame with one GPU),
which can be applied in the real-time context-aware system and assist surgeons in the real-world surgical operating,
including warning generating, process monitoring as well as staff scheduling.
Such real-time notification and online assistance systems have large potentials to become the key component in the modern operating rooms, especially with the gradual development of robotic minimally invasive surgery.
In addition, automatic tool presence detection and phase recognition of surgical videos play the significant roles in some postoperative applications, such as surgical report writing, video database indexing, skill assessment and postoperative review.
The proposed MTRCNet-CL is general enough that not only can analyze the cholecystectomy, but also can be extended to address multiple tasks in other types of surgical videos, such as cataract surgery, robotic surgery, etc.
In essence, there also exist high correlations between tool usage and surgical activity in other surgical videos.
As long as there exist high correlations in the videos, the mapping matrix can be learned and the proposed correlation loss can be leveraged to improve the performance by penalizing the inconsistency.

\section{Conclusion}
In this paper, we present a novel architecture with correlation loss (MTRCNet-CL) to simultaneously detect surgical tool and recognize phase.
Specifically, the designed architecture shares the features in the early layers and holds respective higher layers for corresponding tasks.
LSTM is employed in the phase branch to model sequential dependencies.
More importantly, the correlation loss with learned mapping matrix is proposed to enforce the consistency of predictions of two tasks.
To this end, our framework is able to sufficiently capture the close relatedness by encouraging the interaction of two branches.
Extensive experiments have validated the effectiveness of our method on a large surgical video dataset, outperforming the state-of-the-art methods.

\section*{Acknowledgments}
This work is supported by grants from the National Basic Program of China, 973 Program (Project No. 2015CB351706), the Research Grants Council of HKSAR, China (Project No. 14225616), the Hong Kong Innovation Fund (Project No. ITT/024/17GP) and the Shenzhen Science and Technology Program (Project No. JCYJ20170413162617606).


\bibliography{ref}

\end{document}